\documentclass[11pt]{article}

\usepackage[preprint]{acl}

\usepackage{times}
\usepackage{latexsym}

\usepackage[T1]{fontenc}
\usepackage[T5]{fontenc}
\usepackage[utf8]{inputenc}
\usepackage{microtype}
\usepackage{inconsolata}
\usepackage{graphicx}

\title{Credit C-GPT: A Domain-Specialized Large Language Model for Conversational Understanding in Vietnamese Debt Collection}

\author{
Nhung Nguyen Thi Hong \\
\texttt{nhung.nguyen@emandai.net}
\And
Cuong Nguyen Dang \\
\texttt{cuong.nguyen@emandai.net}
\And
Tri Le Ngoc\thanks{Corresponding author.} \\
\texttt{tri.ln@emandai.net}
}

\begin{document}
\maketitle
\begin{abstract}
Debt collection is a critical function within the Banking, Financial Services, and Insurance (BFSI) sector, relying heavily on large-scale, human-to-human conversational interactions conducted primarily in Vietnamese contact centers. These conversations are characterized by informal spoken language, emotional volatility, and complex domain-specific reasoning, posing significant challenges for traditional natural language processing (NLP) systems. In this paper, we introduce Credit C-GPT, a 7-billion-parameter domain-specialized large language model fine-tuned for conversational understanding in Vietnamese debt collection. Credit C-GPT unifies multiple conversational intelligence tasks—including dialogue understanding, sentiment recognition, intent detection, call stage classification, and structured slot-value extraction—within a single reasoning-based model. We describe the data construction, annotation process, and training methodology, and evaluate the model on proprietary human-annotated datasets. Experimental results demonstrate consistent improvements over traditional pipeline-based baselines, suggesting that domain-specialized conversational LLMs offer a scalable and privacy-preserving solution for real-time assistance and post-call analytics in enterprise contact centers.
\end{abstract}

\section{Introduction}

Debt collection contact centers handle millions of customer interactions daily, forming a core operational pillar of the BFSI industry. These interactions are not only transactional but also emotionally charged, involving negotiation, resistance, compliance obligations, and dynamic conversational strategies. Understanding such conversations at scale is essential for improving recovery rates, ensuring regulatory compliance, coaching agents, and enhancing customer experience.

However, existing NLP solutions deployed in contact centers largely rely on modular pipelines composed of automatic speech recognition (ASR), intent classification, sentiment analysis, and slot-filling models \citep{tur2011spoken, Xu2014ContextualDC, Yao2013RecurrentNN, Bhargava2013EasyCI, Ravuri2015RecurrentNN, 6737243, Tr2012TowardsDU, MartnDoas2024SpeechER}. These systems typically operate on short utterances, assume clean text input, and lack the ability to reason over long conversational context. As a result, they struggle with spoken language phenomena such as disfluencies, code-switching, implicit intent, emotional escalation, and stage transitions that are pervasive in debt collection calls.

Recent advances in large language models (LLMs) have demonstrated strong general-purpose reasoning and language understanding capabilities \citep{Wei2022EmergentAO, Brown2020LanguageMA, Achiam2023GPT4TR, Minaee2024LargeLM, Hagos2024RecentAI, Khan2025AdvancesIL}. Inspired by domain-specialized models such as BloombergGPT \citep{Wu2023BloombergGPTAL} for finance, there is growing interest in adapting LLMs to specific enterprise domains where data distributions, terminology, and reasoning patterns differ substantially from open-domain text \cite{Wu2023BloombergGPTAL, Gururangan2020DontSP, Yang2020FinBERTAP}. Nevertheless, the application of LLMs to conversational debt collection intelligence remains underexplored in the research literature.

In this work, we present Credit C-GPT, a 7B-parameter large language model fine-tuned on domain-specific conversational data from Vietnamese debt collection contact centers. Credit C-GPT is trained to perform multiple interrelated conversational analysis tasks within a single model, including:

\begin{itemize}
\item Understanding multi-turn, real-world spoken dialogue between agents and customers;
\item Recognizing customer emotion and attitude;
\item Identifying dynamic intents that evolve across the call;
\item Classifying call stages such as verification, negotiation, commitment, and closure;
\item Extracting structured slot-value information such as promised payment amounts and dates.
\end{itemize}

By unifying these tasks into a single conversational reasoning model, Credit C-GPT reduces system complexity and improves consistency across predictions. The main contributions of this paper are threefold:
\begin{enumerate}
\item We introduce the first domain-specialized conversational LLMs tailored specifically for Vietnamese BFSI debt collection.
\item We propose a unified modeling approach that replaces traditional multi-model NLP pipelines with a single reasoning-based LLMs.
\item We provide an empirical evaluation demonstrating the effectiveness of domain-adaptive fine-tuning for complex, spoken conversational tasks.
\end{enumerate}

The remainder of this paper is organized as follows. Section 2 reviews related work on domain-specific LLMs and conversational analytics. Section 3 describes the model architecture. Section 4 details the dataset and annotation process. Section 5 outlines the training methodology. Section 6 presents experimental results and evaluation. Finally, Section 7 discusses limitations and future work, and Section 8 concludes the paper.

\section{Related Work}

\subsection{Domain-Specialized Language Models}
Recent efforts in training models exclusively on domain-specific data have shown that smaller, specialized models can outperform general-purpose LLMs on in-domain tasks, particularly in scientific \citep{Taylor2022GalacticaAL} and biomedical domains \citep{Bolton2024BioMedLMA2, Luo2022BioGPTGP, Lehman2023DoWS}, as well as in finance with FinBERT \citep{Yang2020FinBERTAP} and BloombergGPT \citep{Wu2023BloombergGPTAL}. While these models demonstrate strong in-domain performance, they are primarily trained on written text, limiting their applicability to spoken, multi-turn conversational settings.

\subsection{Pipeline-based Conversational Analytics}
Research on conversational analytics has traditionally relied on modular NLP pipelines comprising intent classification, sentiment or emotion analysis, and slot-filling components \citep{Bhargava2013EasyCI, Ravuri2015RecurrentNN, Araci2019FinBERTFS}. Although effective in constrained scenarios, these approaches struggle with spoken-language phenomena such as disfluencies, overlapping speech \cite{Teleki2024QuantifyingTI, Lian2024TowardsHS, Faruqui2021RevisitingTB}, long-range conversational dependencies \citep{Varzaneh2024TransformingNW}, and dynamically evolving user intent \citep{Kanani2025TheEO}. More recent work has explored the use of LLMs to jointly model multiple conversational tasks in an end-to-end manner, highlighting the advantages of unified, context-aware modeling \cite{Yi2024ASO}.

\subsection{Unified LLM-based Conversational Modeling}
Despite these advances, most existing domain-specialized conversational LLMs primarily focus on English and other high-resource languages, with limited attention to Vietnamese spoken dialogue in enterprise settings \citep{Van2022ViWOZAM, Dinh2024MultiDialectVT}. In contrast, our work targets Vietnamese debt collection conversations, which present additional challenges such as colloquial expressions, code-switching, and domain-specific pragmatic cues. By adapting a large language model to Vietnamese conversational data and unifying multiple analytical tasks within a single model, our approach addresses a gap in prior research on domain-specific conversational analytics for low- to mid-resource languages.

\section{Model Architecture and Inference}

\subsection{Base Architecture}
Credit C-GPT is based on the Qwen2.5-7B Instruct decoder-only Transformer architecture \citep{Yang2024Qwen25TR}, which employs rotary positional embeddings and multi-head self-attention. We adopt the base architecture without architectural modification, focusing instead on domain adaptation through data curation, supervised instruction tuning, and training strategy. Qwen is selected for its strong multilingual capabilities, robustness in instruction-following tasks, and favorable performance–efficiency trade-offs for enterprise deployment, particularly in long-context conversational settings.

\subsection{Inference Strategy}
The model supports context-aware turn-level inference and can be extended to post-call analytical workflows. In real-time agent assist scenarios, Credit C-GPT performs turn-level inference by conditioning on the entire conversational history in a rolling context and generating structured annotations for a designated target turn. This approach enables the model to leverage long-range conversational context while maintaining low latency and fine-grained predictions required for real-time applications.

For post-call analytics, the model conditions on the full multi-turn transcript as contextual input to generate detailed turn-level annotations across the conversation. These turn-level outputs can subsequently be aggregated at the call level to support downstream tasks such as call summarization, final outcome determination, and the extraction of high-level business metrics. This flexibility enables deployment across multiple operational workflows using a single model checkpoint.

\section{Dataset and Annotation}

\subsection{Data Collection}
The training data for the model is constructed based on simulated conversational scenarios that are carefully recreated from real-world debt collection calls conducted at BFSI contact centers. Through multiple years of hands-on experience supporting the deployment and operation of conversational bots in the BFSI domain, our organization has built a team of domain experts with extensive operational knowledge and deep professional expertise in debt collection.

Leveraging this domain experience, we reconstruct representative conversational scenarios that reflect the full spectrum of real operational situations, including cooperative responses, resistance, negotiation, deferment, and payment commitment. Based on these scenarios, simulated phone calls are conducted between debt collection agents and customers, strictly adhering to real-world business logic, regulatory constraints, and conversational strategies used in production environments.

All audio data generated from the simulated calls is subsequently transcribed into text. During dataset construction, special care is taken to ensure that the transcripts faithfully reflect real-world conversational conditions, including but not limited to:
\begin{itemize}
\item Background noise and acoustic interference commonly observed in contact center environments;
\item Disfluencies such as hesitations, repetitions, and self-corrections;
\item Overlapping speech between agents and customers;
\item Incomplete, fragmented, or semantically underspecified utterances;
\item Informal, verbose, and emotionally charged spoken expressions.
\end{itemize}
Although the dataset is generated through simulation, the scenarios and conversations are designed and executed by experienced domain practitioners, allowing the data to achieve a high level of realism while ensuring strict compliance with data privacy and confidentiality requirements. This approach provides a reliable and scalable data foundation for training conversational language models that must robustly handle spoken-language phenomena in debt collection interactions.

\subsection{Annotation Schema}
Each conversation is segmented into dialogue turns and annotated by trained domain experts following detailed annotation guidelines. The annotation schema includes multiple interdependent labels:
\begin{itemize}
    \item Emotion: capturing the customer’s affective state (e.g., neutral, negative, positive);
    \item Sentiment: indicates the level and form of negative or confrontational behavior expressed in the customer’s utterance. (e.g., insult, refusal, threat);
    \item Intent: identifying the underlying goal or response strategy expressed by the customer, which may evolve throughout the call;
    \item Call Stage: labeling high-level phases of the debt collection process such as opening, verification, negotiation, commitment, and closure;
    \item Slot-Value Pairs: extracting structured entities including payment amount, promised date, days past due, customer name.
\end{itemize}
Annotations are performed at both turn-level and segment-level to capture dynamic transitions in intent and call stage. Inter-annotator agreement is monitored throughout the labeling process to ensure consistency and reliability. The final dataset is partitioned into training, validation, and test splits with no overlap at the call level, preventing information leakage across splits.

In contrast to traditional intent classification datasets that treat utterances independently, this dataset preserves full conversational context, enabling Credit C-GPT to learn long-range dependencies and domain-specific reasoning patterns inherent to debt collection dialogues.

\subsection{Dataset Statistics}
The full dataset consists of 17,000 debt collection conversations, comprising approximately 336,935 dialogue turns. The data is split at the conversation level into training and validation sets using a 90/10 ratio to prevent information leakage across splits.

In addition, a held-out real-world test set is constructed from production Vietnamese debt collection calls that are not used during training or validation. This test set serves to evaluate the model’s generalization ability from simulated to real conversational data.

\begin{table}[t]
\centering
\caption{\textbf{Dataset statistics.}}
\label{tab:dataset}
\begin{tabular}{lrrr}
\hline
Split & Conversations & Turns & Turns/Conv \\
\hline
Train & 15,300 & 298,755 & 19.5 \\
Valid & 1,700  & 38,160  & 22.4 \\
Test  & 400    & 11,955  & 29.9 \\
\hline
\end{tabular}
\end{table}

Compared to the simulated training data, real-world test conversations are notably longer and exhibit higher variability in conversational flow and emotional dynamics, reflecting the complexity of authentic debt collection interactions in production environments.

\section{Training Methodology}

Credit C-GPT is adapted to the debt collection domain through supervised instruction tuning on annotated conversational data. Training samples are constructed as instruction–input–output triples, where the instruction specifies the required analytical tasks, the input consists of a dialogue segment or a full transcript together with a designated target turn, and the output is a structured JSON representation of the extracted information.

Rather than introducing task-specific heads, Credit C-GPT is trained as a unified generative model that jointly predicts sentiment, intent, call stage, and slot–value information in a single forward pass, enabling consistent cross-task reasoning. Domain instructions and output schemas are embedded directly into the training prompts, allowing the model to internalize task boundaries without architectural modifications.

To improve training efficiency, we apply QLoRA (Quantized Low-Rank Adaptation) \citep{Dettmers2023QLoRAEF} during fine-tuning. The base model is quantized to 4-bit precision, while low-rank adaptation parameters are trained on selected Transformer layers, substantially reducing memory and compute requirements while preserving inference performance and domain adaptability.

Training batches include a mixture of short dialogue excerpts and long multi-turn conversations to enhance robustness to varying context lengths. Hyperparameters are selected based on validation performance, and training is conducted on a GPU cluster consisting of four NVIDIA L40S GPUs, which is suitable for fine-tuning models at the 7B parameter scale. No external financial corpora are used beyond the proprietary dataset.

\section{Evaluation}

\subsection{Evaluation Tasks and Metrics}
We evaluate Credit C-GPT across five core conversational intelligence tasks: emotion recognition, sentiment classification, intent detection, call stage classification, and slot–value extraction. For all tasks, the model operates in a turn-level inference setting, where each target turn is evaluated conditioned on the full preceding conversational history.

At inference time, the input consists of the dialogue context and a designated target turn, formatted according to the instruction schema used during training. The model produces structured JSON outputs containing task-specific predictions, including categorical labels and extracted slot–value pairs.

Classification tasks are evaluated using accuracy. Slot–value extraction performance is measured using entity-level accuracy.

Evaluation is conducted on a held-out test set consisting of 400 real-world Vietnamese debt collection calls, comprising approximately 11,955 annotated dialogue turns. These conversations reflect authentic contact center interactions and are not used during training or validation.

In addition to automatic metrics, a subset of model outputs is reviewed by domain experts to assess qualitative correctness and practical usability in enterprise workflows.

\subsection{Baselines}

We compare Credit C-GPT against a BERT-based encoder model implemented as a standard classification pipeline. The baseline is trained on the same domain-specific Vietnamese conversational dataset as Credit C-GPT and uses identical train--validation--test splits.

The BERT-based pipeline consists of multiple independently trained components commonly deployed in contact center analytics, including:
\begin{itemize}
\item Utterance-level intent classifiers based on fine-tuned transformer encoders;
\item Sequence labeling models for slot extraction;
\item Standalone sentiment analysis models for emotion detection.
\end{itemize}
These baselines operate independently and do not share conversational context beyond a limited window, reflecting typical production deployments.

We additionally compare against GPT-5, prompted to perform the same annotation tasks. GPT-5 and Credit C-GPT receive identical conversational inputs and are evaluated using the same output schema and metrics, enabling a direct comparison between domain-specialized fine-tuning and prompt-based adaptation of general-purpose LLMs.

\subsection{Results and Analysis}
Table 2 reports the performance on core conversational classification tasks, including emotion, sentiment, intent, and call stage classification. Credit C-GPT consistently outperforms traditional BERT-based pipeline models across all tasks, with particularly strong improvements in intent detection and call stage classification. These results highlight the advantage of unified conversational modeling over modular pipelines, which often suffer from error propagation and limited context utilization.

While general-purpose LLMs such as GPT-5 achieve higher absolute scores, Credit C-GPT demonstrates competitive performance despite being substantially smaller and trained exclusively on domain-specific Vietnamese conversational data. Notably, Credit C-GPT shows strong robustness on tasks requiring long-context reasoning and temporal modeling of dialogue progression.

To further isolate the impact of domain-specific fine-tuning from model architecture and scale, we additionally compare Credit C-GPT with the base Qwen2.5 7B Instruct model using prompt-based adaptation without domain-specific training. As both models share the same backbone and parameter scale, this comparison provides a controlled evaluation of domain adaptation.

As shown in Tables~\ref{tab:performance} and~\ref{tab:extraction}, Credit C-GPT consistently outperforms Qwen across all tasks, with particularly large gains in intent detection, call stage classification, and debt-related slot–value extraction. While Qwen exhibits strong general instruction-following capability, it often struggles with implicit intent shifts, colloquial Vietnamese expressions, and domain-specific negotiation patterns common in debt collection conversations. This further suggests that domain supervision, rather than model scale alone, plays a critical role in complex multi-turn conversational understanding.

Beyond LLM-based approaches, encoder-based models for Vietnamese such as PhoBERT \citep{Nguyen2020PhoBERTPL} can be applied to slot filling in a token classification setup. However, such approaches typically operate on single utterances or short context windows and struggle to capture long-range dependencies across multi-turn conversations, which are critical for extracting commitments and temporal information in debt collection calls.

\begin{table}[t]
\centering
% \small
\caption{\textbf{Performance Comparison on Conversational Classification Tasks.}}
\label{tab:performance}
\resizebox{\columnwidth}{!}{%
\begin{tabular}{lcccc}
\hline
Task & BERT-based$_{\uparrow}$ & Qwen2.5 7B Instruct$_{\uparrow}$ & \textbf{Credit C-GPT}$_{\uparrow}$ & GPT-5$_{\uparrow}$ \\
\hline
Emotion & 0.74 & 0.78 & 0.90 & \textbf{0.97} \\
Sentiment & 0.70  & 0.70 & 0.89  & \textbf{0.95} \\
Intent  & 0.65   & 0.59  & 0.77  & \textbf{0.84} \\
Call Stage & 0.72  & 0.62  & 0.88  & \textbf{0.92} \\
\hline
All Tasks(avg) & 0.73  & 0.67  & 0.86  & \textbf{0.92} \\
\hline
\end{tabular}}
\end{table}

GPT-5 is evaluated as a general-purpose externally hosted LLM via prompting without domain-specific fine-tuning.

Table 3 presents entity-level Accuracy for slot–value extraction. Credit C-GPT achieves strong performance across key business-critical entities, particularly for agent and customer identifiers as well as debt-related attributes. Although GPT-5 attains higher absolute scores, Credit C-GPT remains competitive given its smaller scale and domain-restricted training data.

\begin{table}[t]
\centering
\caption{\textbf{Slot–Value Extraction Performance (Entity-level Accuracy).}}
\label{tab:extraction}
\resizebox{\columnwidth}{!}{%
\begin{tabular}{lccc}
\hline
Slot & Qwen2.5 7B Instruct$_{\uparrow}$ & \textbf{Credit C-GPT}$_{\uparrow}$ & GPT-5$_{\uparrow}$ \\
\hline
agent\_name & 0.73 & 0.93 & \textbf{0.98} \\
customer\_name & 0.70 & 0.85  & \textbf{0.92} \\
total\_debt  & 0.78 & 0.90    & \textbf{0.97}\\
days\_past\_due  & 0.62 & 0.89    & \textbf{0.97} \\
promised\_payment\_date  & 0.65 & 0.70    & \textbf{0.83}\\
promised\_payment\_amount  & 0.59 & 0.72    & \textbf{0.84}\\
due\_date  & 0.60 & 0.75    & \textbf{0.83}\\
\hline
\end{tabular}}
\end{table}

Although Credit C-GPT does not surpass GPT-5 in absolute metrics, it offers a favorable balance between performance, domain alignment, and deployability. Credit C-GPT is designed to satisfy the stringent security, privacy, and data governance requirements of BFSI environments, enabling on-premise deployment, fine-tuning on proprietary data, and controlled structured outputs. In practical debt collection workflows, these constraints often outweigh marginal gains in raw accuracy, making Credit C-GPT a competitive and operationally viable solution.

Qualitative analysis further indicates that Credit C-GPT effectively captures implicit intent shifts, emotional escalation, and negotiation dynamics that are difficult to model using traditional pipeline-based systems. Together, these findings support the effectiveness of domain-specialized LLMs for real-world conversational analytics in Vietnamese debt collection contact centers.

\section{Discussion and Limitations}

While Credit C-GPT demonstrates strong performance on conversational debt collection analysis, several limitations should be acknowledged. First, the model is trained primarily on proprietary domain-specific data, which limits direct reproducibility and comparison by the research community. This constraint is common in enterprise NLP settings but nonetheless restricts open benchmarking and independent validation.

Second, similar to other large language models, Credit C-GPT remains sensitive to prompt formulation and may occasionally produce hallucinated outputs in low-information or ambiguous conversational contexts. Although structured output schemas mitigate this behavior to some extent, complete elimination of such errors remains an open challenge.

From a deployment perspective, Credit C-GPT is designed for on-premise enterprise environments and is served in a 4-bit quantized configuration on a single GPU. While vertical scaling can improve throughput and latency, a comprehensive system-level performance and cost benchmark is beyond the scope of this paper.

Due to the absence of publicly available Vietnamese conversational datasets in the BFSI domain, evaluation is conducted exclusively on proprietary, real-world test data. Consequently, we do not assess potential catastrophic forgetting on general-domain benchmarks, which we leave for future investigation.

Future work will explore reinforcement learning from human feedback, tighter output validation constraints, and multilingual expansion to additional regional markets. Further investigation into explainability and regulatory compliance is also warranted given the high-stakes nature of BFSI applications.

\section{Conclusion}

This paper introduces Credit C-GPT, a 7B-parameter domain-specialized large language model designed for understanding spoken Vietnamese conversations in BFSI debt collection contact centers. By unifying emotion recognition, intent detection, call stage classification, and slot-value extraction within a single reasoning model, Credit C-GPT addresses key limitations of traditional NLP pipelines when applied to informal, multi-turn, and emotionally dynamic spoken dialogue.

Beyond accuracy improvements, Credit C-GPT is designed to operate as an on-premise analytical model, reflecting the strict data security and privacy requirements of BFSI environments. The model is positioned as a drop-in replacement for conventional NLP components in contact center analytics, enabling improved performance on Vietnamese spoken conversations while remaining non-generative with respect to customer-facing responses. As a result, the risk of uncontrolled language generation and hallucination is substantially reduced compared to end-to-end generative dialogue systems.

Our empirical results highlight the effectiveness of domain-adaptive fine-tuning for spoken Vietnamese conversational data and suggest that specialized conversational LLMs offer a practical and reliable path toward modernizing enterprise contact center intelligence under real-world deployment constraints.

\bibliography{anthology}

\end{document}